\documentclass[conference]{IEEEtran}
\IEEEoverridecommandlockouts

\IEEEpubid{\makebox[\columnwidth]{978-1-7281-6251-5/20/\$31.00~\copyright~2020 IEEE \hfill} \hspace{\columnsep}\makebox[\columnwidth]{ }}

\usepackage{cite}
\usepackage{amsmath,amssymb,amsfonts}
\usepackage{algorithmic}
\usepackage{graphicx}
\usepackage{textcomp}
\usepackage{xcolor}
\usepackage[font=footnotesize,caption=false]{subfig}
\usepackage{balance}
\def\BibTeX{{\rm B\kern-.05em{\sc i\kern-.025em b}\kern-.08em
    T\kern-.1667em\lower.7ex\hbox{E}\kern-.125emX}}
\begin{document}

\title{Knowledge- and Data-driven Services for Energy \\ 
Systems  using Graph Neural Networks
\thanks{This research has received funding from the European
Research Council under the European Unions Horizon 2020
research and innovation programme (grant agreement no.
731232).}
}

\author{\IEEEauthorblockN{Francesco Fusco, Bradley Eck,
Robert Gormally, Mark Purcell, Seshu Tirupathi}
\IEEEauthorblockA{IBM Research Europe\\
Dublin, Ireland\\
Email: francfus@ie.ibm.com}
}

\maketitle

\IEEEpubidadjcol

\begin{abstract}
The transition away from carbon-based energy sources poses several challenges for the operation of electricity distribution systems. 
Increasing shares of distributed energy resources (e.g. renewable energy generators, electric vehicles) and internet-connected sensing and control devices (e.g. smart heating and cooling) require new tools to support accurate, data-driven decision making. Modelling the effect of such growing complexity in the electrical grid is possible in principle using state-of-the-art power-power flow models. In practice, the detailed information needed for these physical simulations may be unknown or prohibitively expensive to obtain. Hence, data-driven approaches to power systems modelling, including feed-forward neural networks and auto-encoders, have been studied to leverage the increasing availability of sensor data, but have seen limited practical adoption due to lack of transparency and inefficiencies on large-scale problems.
Our  work addresses this gap by proposing a data- and knowledge-driven probabilistic graphical
model for energy systems  based on the framework of graph neural networks (GNNs). The model can explicitly factor in domain knowledge, in the form of grid topology or physics constraints, thus resulting in sparser architectures and much smaller parameters dimensionality when compared with traditional machine-learning models with similar accuracy. 
Results obtained from a real-world smart-grid demonstration project show how the GNN was used to inform grid congestion predictions and market bidding services for a distribution system operator participating in an energy flexibility market.

\end{abstract}


%
\IEEEpeerreviewmaketitle

\section{Introduction} \label{sec:intro}

The complexity of the electrical energy system is increasing with higher numbers of distributed devices for energy consumption, generation and storage. As an example, between 2007 and 2017, the European installed capacity of renewable energy almost doubled from 258 GW to 512 GW, primarily coming from new photovoltaic and wind generators \cite{Child2019}. 
The centralised architecture of electricity dispatch, generation and distribution is transitioning toward a more distributed orchestration of control, optimization and market services for the balancing of supply and demand. The Internet of Things (IoT) technology has seen massive deployment of inter-connected sensing and control devices through residential, commercial and industrial settings. 
The growing volume and heterogeneity of data enables novel data services to support the need for automated, data-driven decision making that is ultimately required to coordinate such increasingly complex energy system. 

Modelling the electrical grid is the foundation of many of the required smart data services for energy utilities. An energy system model provides for a digital, mathematical representation of the collective behavior of the complex set of interconnected assets, from consumers and generators up to distribution transformers and electrical substations through power lines and power-flow control devises. Traditionally, this model is obtained from physics-based sets of power flow equations \cite{Abur2004} and informs several processes. Predictions (simulations) of the impact on the grid assets of observed and predicted (or desired) energy consumption and generation profiles can be assessed. Furthermore, the full state of the grid, that is the value of all electrical quantities at the grid assets, can be inferred from any available sensor data through a model inversion process known as state estimation \cite{Monticelli2000}. Asset and sensor data anomalies can also be diagnosed based on statistical analysis of the deviation between the model inference and the sensor observations, a process known as residual analysis \cite{Abur2004}. 

The increasing complexity of energy systems, however, makes the derivation of a sufficiently detailed and accurate mathematical model from first principles, with well-defined values for its electrical parameters, an often prohibitive task. The issue is particularly evident for medium- to low-voltage distribution grids, where the growing need for accurate modelling is more and more critical and, at the same time, up-to-date physical models and detailed electrical parameters are often either not available or inaccurate. At the same time, the increasing availability of sensor data from smart metering infrastructure or IoT devices, makes data-driven modelling approaches based on machine learning more appealing. Among others, solutions based on feed-forward neural networks and auto-encoders have been studied to solve power systems state-estimation and prediction \cite{Barbeiro2014,Zamzam2019}. Not accounting for the domain knowledge often accompanying the sensor-data, in the form of grid connectivity of physics constraints, is however a strong limiting factor in the accuracy, transparency and, ultimately, practical applicability of these methods \cite{Zhang2019}. Knowledge-based systems have recently emerged to enrich IoT sensor data with a semantic reasoning layer that supports explicitly embedding domain knowledge into several machine learning tasks such as data exploration, feature engineering and model creation\cite{Schumilin2017,Zhang2017,Eck2019,Chen2018,Ploennigs2017}.
A promising research direction to create data-driven models for power systems that can be explicitly informed from knowledge-based systems is the use of probabilistic graphs \cite{Hu2011,Cosovic2016a,Fusco2018}. Probabilistic graphical models provide for a natural way to represent structural relationships between the variables of a complex system based on domain knowledge such as electrical grid connectivity, known localised correlation structure or even physical equations. In order to overcome limitations of the existing approaches, requiring costly belief propagation algorithms often designed based on ad-hoc heuristics to solve the inference problem for specific structures of the graph or functional form of the probability densities, a fully data-driven approach is proposed here, based on the recently emerged framework of graph neural networks (GNNs) \cite{Scarselli2009,Casas2020}. The proposed GNN model was developed in order to enable congestion management and market bidding data services for a distribution system operator (DSO) participating in energy flexibility markets.

After an overview, in Section \ref{sec:architecture}, of the data-analytic services that were developed in the context a smart-grid research project\cite{goflex}, Section \ref{sec:model} details the proposed modelling approach based on GNNs, underlying these services. Results from experiments during the live deployment of the data services at a pilot demonstration of the flexibility market are then discussed in Section \ref{sec:results}. Conclusions and final remarks are given in Section \ref{sec:conclusion}.

\section{Data Services for Energy Flexibility markets} \label{sec:architecture}

A set of smart-grid technologies to enable an energy flexibility market was developed within the context of  the research project GOFLEX, funded by the European Union and involving a consortium of energy utilities,
technology providers and research institutions across Europe \cite{goflex,Eck2019eenergy}.
The flexibility market enabled residential or industrial electrical prosumers (consumers and producers) to actively participate in the energy system by offering to sell the flexibility in their energy production and/or consumption processes. Distribution system operators (DSOs) offered to buy the available flexibility from the market in order to solve potential issues that might be caused by excessive distributed renewable energy generation, thus increasing the grid capacity margin without the need for expensive capital investments.

\begin{figure}
\centering
\includegraphics[width=0.46\textwidth]{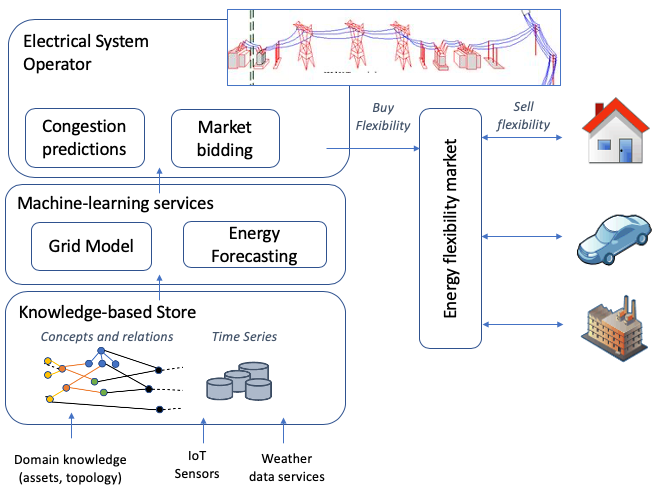}
\caption{Energy data services based on knowledge- and data-driven machine-learning in the context of energy flexibility markets. }
\label{fig:data-services}
\end{figure}

A set of specific data-analytic services, as shown in Fig. \ref{fig:data-services}, were developed in order to enable the participation of energy utilities, such as DSOs, in the flexibility markets. Grid congestion prediction services estimate the future behavior, over a 24-48 hours window, of a number of electrical quantities of interest at important grid assets, along with the likelihood of them operating outside some user-defined threshold. Where instances of congestions are flagged, market bidding services estimate the corresponding amount of energy flexibility required at different points of the grid in order to avoid the predicted congestions. These specific use-cases of grid prediction and bidding data services are built on top of core machine-learning functionalities, including a model of the electrical grid as well as distributed energy forecasting models of demand and renewable generation. External data services, such as high-resolution weather predictions and grid sensor data from IoT sensors or traditional SCADA systems ultimately drive the data services. The sensor data are enriched by the semantic layer of a knowledge-based system that incorporates available domain expertise in the form, for example, of known relations between grid asset, variables and the sensors.  

A cloud-based architecture was designed to enable the data services of Fig. \ref{fig:data-services}, in the context of the smart-grid demonstration pilots of the project in \cite{goflex}. Following a micro-services design pattern, the knowledge-based time-series management micro-service was hosted on a combination of relational and graph database cloud storage services. The machine-learning modelling services for energy forecasting and grid modelling relied on a combination of containerised (orchestration and time-consuming tasks such as model training) and serverless (frequent, bursty workloads such as computing model predictions) cloud-computing infrastructure . Data ingestion and communication, both internally and with external consumers through high-level data services (e.g. the congestion predictions and market bidding) relied on asynchronous messaging based on Message Queuing Telemetry Transport (MQTT) and Advanced Message Queuing Protocol (AMQP) and on serverless computing. Further architecture details were provided in \cite{Eck2019,Chen2018}. 

The benefits of knowledge-based time-series systems for managing and automating several aspects of the deployment of energy forecasting machine-learning models in large industrial-scale deployments were detailed in \cite{Eck2019}. In the following, the paper focuses on the specific development of the grid modelling services, which were based on a novel machine-learning model of the electrical grid that can factor in both the sensor data as well as the available domain knowledge, specifically in the form of power network connectivity, by leveraging the graph neural network framework.

\section{Proposed Graph neural network model} \label{sec:model}

A power systems model can be generally expressed as \cite{Abur2004}:
\begin{equation}
y_t = f(x_t) + \varepsilon_t ,
\label{eq:powerflow}
\end{equation}
where: $y_t$ is a set of electrical quantities of interest at a given time $t$, such as active/reactive power, current magnitude, voltage magnitude and angles, etc.; $x_t$ is the state variable and it denotes the minimum independent set of variables that fully describe the electrical system (a typical choice for the state variables in physics-based models is the set of voltage magnitudes and angles at all nodes of the grid, but it might be considered as a latent variable in a machine-learning-based model); $f(\cdot)$ is the set of, generally non-linear, power-flow equation relating the state variable to all electrical quantities of interest; $\varepsilon_t$ is an error term that quantifies modelling uncertainty or sensor noise. 

By taking into account the structure of the energy system, the joint density of the model variables in (\ref{eq:powerflow}), can be factorized as follows:
\begin{equation}
p(y,x) = \prod_{i=1}^n \phi_i (y_i | x_i) \prod_{j=1}^{\mathcal{N}(i)} \psi_{ij} (x_i,x_j),
\label{eq:factorized}
\end{equation}
where $y_i$, $x_i$, for $i=1,\ldots n$, are $n$ subsets of the system and state variables, $\phi_i(\cdot)$ are conditional densities and $\psi_{ij}(\cdot)$ are joint densities between each state variable $x_i$ and its neighbours $\mathcal{N}(i)$. 

Under Gaussian assumption, the densities $\phi(\cdot)$ and $\psi(\cdot)$ can be further specified as: 
\begin{equation}
\begin{aligned}
&\phi_i(y_i|x_i) \propto  exp\left(\frac{1}{2}\left[y_i-g_i(x_i)\right]^T\Sigma_i\left[y_i-g_i(x_i)\right]\right) \\
&\psi(x_i,x_j) \propto exp\left(\frac{1}{2}\left[x_{ij}-h_{ij}(x_i,x_j)\right]\Omega_{ij}\left[x_{ij}-h_{ij}(x_i,x_j)\right]\right),
\end{aligned}
\label{eq:factorized_gaussian}
\end{equation}
where $g_i(x_i)$ and $\Sigma_i$ denote, respectively, the conditional mean and covariance of $y_i$ with respect to $x_i$, while $h_{ij}(x_i,x_j)$ and $\Omega_{ij}$ denote the mean and covariance of the combined state variable  $x_{ij}^T=[x_i^T; x_j^T]$.   

The factorization in (\ref{eq:factorized}) effectively defines a probabilistic graph where the nodes are the variables $y_i$, $x_i$ and the edges are defined by both the conditional (direct edges) and joint densities (indirect edges) \cite{Koller2009}. In the specific case of energy grids, the structure of the graph can reflect knowledge about the physical network connectivity but also known localised correlation structure between the system variables. First-principle physical equations from (\ref{eq:powerflow}) can also be incorporated into the graphical model, for example as conditional mean function $g_i(x_i)$ of the Gaussian density in (\ref{eq:factorized_gaussian}).
  
Solving prediction, simulation or inversion problems on the graphical model (\ref{eq:factorized})-(\ref{eq:factorized_gaussian}) involves solving the inference problem based respectively on predictions, assumptions or observations of $y_i$. Inference algorithms on such graphical models take the form of message-passing belief propagation, where information is iteratively exchanged between the nodes across the edges of the graph. In the particular case, considered here, of Gaussian graphical models, message passing takes the form of the sum-product algorithm where updates involve matrix summations and multiplications propagating knowledge about mean and covariance of the distributions. Different forms of the sum-product algorithms have been proposed, depending on the specific format of the conditional or joint densities and even on the structure of the graph  \cite{Cosovic2016a,Fusco2018}. By adopting the most general representation based on factor graphs \cite{Loelinger2007, Koller2009}, the probabilistic inference of physics-based power flow equations using Gaussian belief propagation \cite{Hu2011,Cosovic2016a} as well as the data-driven learning of localised relationships using neural network as nodes in the graph \cite{Fusco2018} has been investigated. Although Gaussian belief propagation, in the form of the sum-product message passing algorithms, solves probabilistic inference more efficiently than centralised problems not accounting for the graph structure \cite{Fusco2018}, it cannot be completely parallelised and might require expensive iterations for solving loopy graphs \cite{Cosovic2016a}. Furthermore, convergence of the belief propagation in the presence of non linear nodes and loops in the graph is not guaranteed and might require ad-hoc heuristic adjustments to the sum-product algorithm, which might depend on the specific structure of the model \cite{Cosovic2016a}.  


\subsection{Message-passing graph neural networks} \label{sec:model_mpnn}

The general framework of graph neural networks (GNNs) incorporates graph structured information in a supervised machine learning model by encoding topological relationships among the nodes of the graph \cite{Scarselli2009}. In particular, each node encodes information about some concept, which is ultimately defined by its features and related concepts, whereby the relationships are encoded by the edges of the graph.   By defining, for each node $k$ of the graph, a state vector, $x_k \in \mathbb{R}^p$, observations about a concept and its features, $y_k \in \mathbb{R}^q$, and an output of interest, $o_k \in \mathbb{R}^r$, a GNN model can be generally written as \cite{Scarselli2009}:
\begin{align}
x_k &= f_k(y_k,x_{\mathcal{N}(k)},y_{\mathcal{N}(k)}) \label{eq:gnn_local_transition}\\
o_k &= g_k(x_k,y_k), \label{eq:gnn_local_output}
\end{align} 
for $k=1,\ldots n$ nodes in the graph. Based on (\ref{eq:gnn_local_transition}), the state vector at a node is related to the data available at the node itself and to both the state and data at neighbour nodes $\mathcal{N}(k)$ through a local transition function $f_k(\cdot)$. In (\ref{eq:gnn_local_output}), the desired target at a node is related to both the data and state vectors at the same node through a local ourput function $g_k(\cdot)$.  

Among the many possibilities of further specifying the GNN model in (\ref{eq:gnn_local_transition})-(\ref{eq:gnn_local_output}), as reviewed in \cite{Battaglia2018}, message passing neural networks (MPNN) provide a suitable framework to solve inference and learning problems on the probabilistic graph in (\ref{eq:factorized_gaussian}) such to emulate principled belief-propagation algorithms \cite{Gilmer2017}. In a MPNN, the local transition function in (\ref{eq:gnn_local_transition}) is expressed as the following iteration: 
\begin{align}
x_k^{(0)} &= f_k^i(y_k) \label{eq:gnn_encode} \\
m_{kj}^{(t)} &= f_{kj}^e (x_k^{(t)},x_j^{(t)}) \quad \forall j \in\mathcal{N}(k) \label{eq:gnn_edge} \\
x_k^{(t+1)} &= f_k^n (m_{kj}, \forall j \in \mathcal{N}(k)) \label{eq:gnn_agg},
\end{align}
where $f_k^i(y_k)$ is a node encoding function, $m_{kj}^{(t)}$ is the message, at iteration step $t$, exchanging information between nodes $k$, $j$ through the edge functions $f_{kj}^e(\cdot)$, and $f_k^n(\cdot)$ is a node aggregation function that updates the state vector at node $k$ based on all incoming messages from neighbouring nodes. The iterative scheme in (\ref{eq:gnn_edge})-(\ref{eq:gnn_agg}) defines a finite-steps message passing algorithm that propagates information through the graph neural network. Note that with respect to the general expression in (\ref{eq:gnn_local_transition}), the explicit dependency of the message passing algorithm on the node concept and features $y_k$ was dropped. It is assumed, in (\ref{eq:gnn_encode}), that the state vector fully encodes the information available in the observations and features available at the node, through the encoding function $f_k^i(\cdot)$. By implementing the sequence of encoding step (\ref{eq:gnn_encode}) and message passing steps (\ref{eq:gnn_edge})-(\ref{eq:gnn_agg}) with neural networks one can obtain a fully differentiable function that can be conveniently trained through back propagation and can offer arbitrarily high capacity of functional approximation.

\subsection{Proposed architecture} \label{sec:model_specific}

The Gaussian graphical model for a networked energy system, defined in (\ref{eq:factorized_gaussian}), is expressed here within the MPNN model framework outlined in Section \ref{sec:model_mpnn}, such that both the non-linear conditional and joint density functions, as well as the belief propagation algorithm, can be jointly modelled with neural networks and trained from standard back propagation. 

\begin{figure}
\centering
\includegraphics[width=0.5\textwidth]{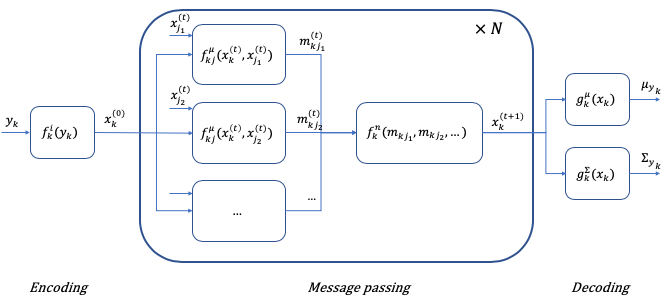}
\caption{Inference in the proposed MPNN, based on (\ref{eq:gnn_encode}) through to (\ref{eq:my_encoder}).}
\label{fig:mpnn}
\end{figure}

The local output functions in (\ref{eq:gnn_local_output}) are used to express both the conditional mean, $\mu_{y_k}$, and variance, $\Sigma_{y_k}$, of the variables at a node $k$ of the graph, namely $y_k$, as function of the state vector at the node: 
\begin{equation}
\begin{aligned}
\mu_{y_k} &= g^\mu_k(x_k) \\
\Sigma_{y_k} &=g^\Sigma_k(x_k),
\end{aligned}
\label{eq:my_decoder}
\end{equation} 
where $g^\mu_k$ and $g^\Sigma_k$ are both fully differentiable parametric functions implemented as neural networks. In (\ref{eq:my_decoder}), only the variances of the individual components of $y_k$ are modelled with the underlying assumption that its covariance matrix is purely diagonal. Note that the output function as written in (\ref{eq:my_decoder}) can also be interpreted as a \emph{decoding} function mapping the internal state representation space to the target space. Accordingly, the node \emph{encoding} function in (\ref{eq:gnn_encode}) can be more specifically written as: 
\begin{equation}
x_k^{(0)} = f_k^i (\mu_{y_k},\Sigma_{y_k}),
\label{eq:my_encoder}
\end{equation}
which provides for the initial iteration of the message passing algorithm in (\ref{eq:gnn_edge})-(\ref{eq:gnn_agg}). 

Figure \ref{fig:mpnn} visualises the overall inference algorithm in the proposed MPNN model as a chain of neural network functions $f_k^i$, $f_{kj}^e$, $f_k^n$,$g_k^\mu$, $g_k^\Sigma$. Note that no iterative algorithm is required, as opposed to principled belief propagation, and only one feedforward pass through the chain generates the required inference estimate. The parameters of the individual functions can be trained by minimising, with any gradient-based algorithm, the following negative log-likelihood objective: 
\begin{equation}
\mathcal{L} \propto \sum_{t=1}^T \sum_{k=1}^N \dfrac{1}{2}\log|\Sigma_{y_k}^t| + \dfrac{1}{2}(y_k^t-\mu_{y_k}^t)^\top \Sigma_{y_k}^{t^{-1}}(y_k^t-\mu_{y_k}^t),
\label{eq:nll}
\end{equation} 
based on the data, $y_k^t$, available at all nodes $k$ an training samples $t$.

Section \ref{sec:model_missingdata} further specifies how the model can be applied in the context of the energy data services described in Section \ref{sec:architecture} in the form of data imputation problems. 

\subsection{Handling missing data} \label{sec:model_missingdata}

The modelling approach defined by (\ref{eq:my_decoder}) and (\ref{eq:my_encoder}) can be thought of as graph-structured variational auto-encoder \cite{Boquet2019}, where mean and covariance of the variables of the energy network are modelled as function of a lower-dimensional latent variable. The proposed model structure provides for a general purpose modelling tool that can be used to solve the problems of prediction or state estimation that are typical in the context of electrical power systems, as reviewed above in section \ref{sec:model}, by framing them as missing data imputation problems. 

In the specific application proposed in section \ref{sec:results}, the same model serves two different data services. Firstly, the model is used to predict the voltage magnitude at some nodes of the energy system based on predictions about the energy consumption and generation available at some other nodes. In the second use-case, the model is also used to determine the energy consumption or generation variation required at some nodes of the network in order to maintain the voltage at certain nodes at a desired reference.  In both cases, the desired prediction or estimation can be obtained by running model inference where only a subset of the data at  the variable nodes is assumed known. From a random initialization of the unknown variables, the sequence of encoding, message passing and decoding steps will provide an updated value for all variables. This imputation procedure can be iterated until convergence simulating a Markov chain, which has been shown that converges to the true marginal distribution of missing values given observed values \cite{Boquet2019}. Similarly, the same model can provide for the basis of other typical problems in energy systems such as state estimation of optimal power flow \cite{Barbeiro2014,Zamzam2019}.

\section{Results} \label{sec:results}

The GNN model outlined in section \ref{sec:model} was developed to inform data-analytics services, such as congestion prediction and market bidding, for electrical system operators trading in energy flexibility markets, as overviewed in Section \ref{sec:architecture}.
In the following, the results obtained from a live deployment of the proposed data services at a real-world demonstration site are discussed. Section \ref{sec:results_data} details the available data and the modelling problem, with the chosen GNN architecture and alternative benchmarks. The validation of the model on the grid voltage prediction task are then detailed in section \ref{sec:results_voltage}. An example of the application of the same grid model to predict grid congestions and generate market bids for energy flexibility is then discussed in section \ref{sec:results_bidding}.

\subsection{Available Data and Model Architecture} \label{sec:results_data}

\begin{figure}
\centering
\includegraphics[width=0.5\textwidth]{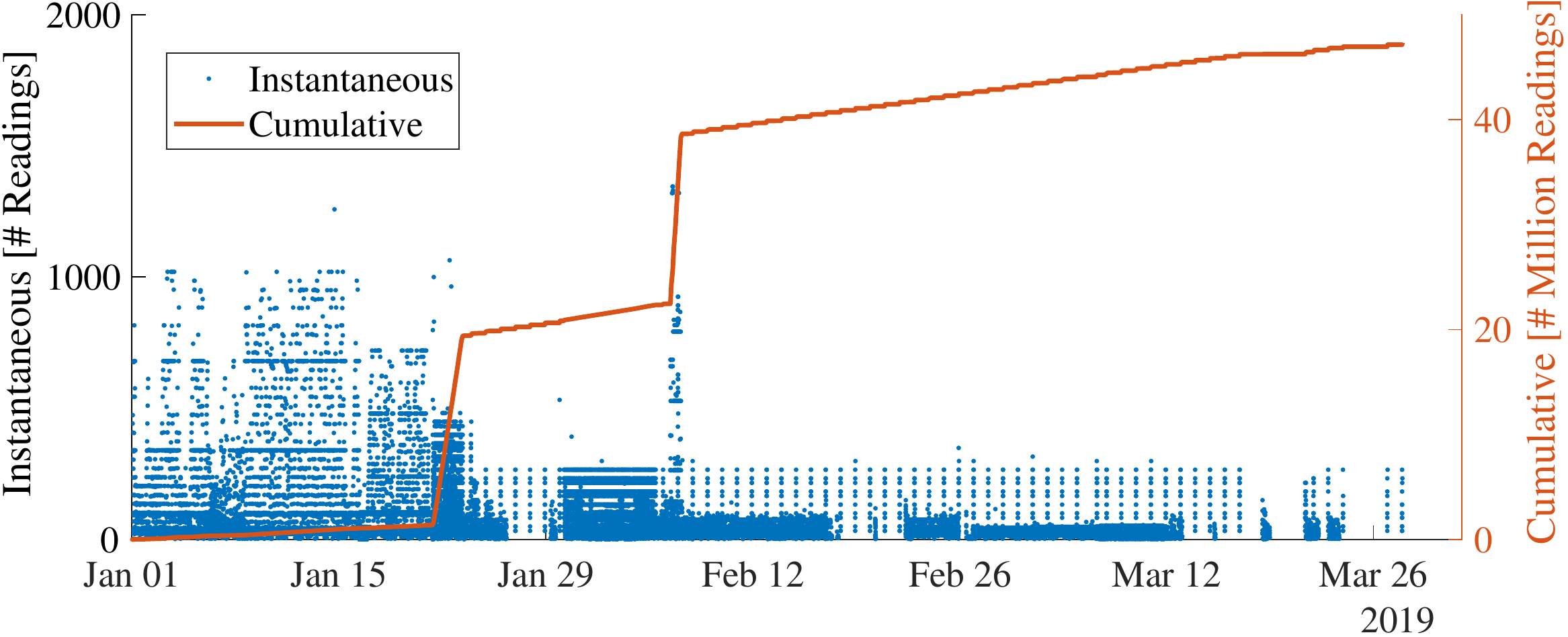}
\caption{IoT sensor data ingestion \cite{Eck2019}.}
\label{fig:data-ingestion}
\end{figure}

\begin{figure}
\centering
      \subfloat[Voltage magnitude observed at a three-phase prosumer. \label{fig:sample-volt}]{%
       \includegraphics[width=0.5\textwidth]{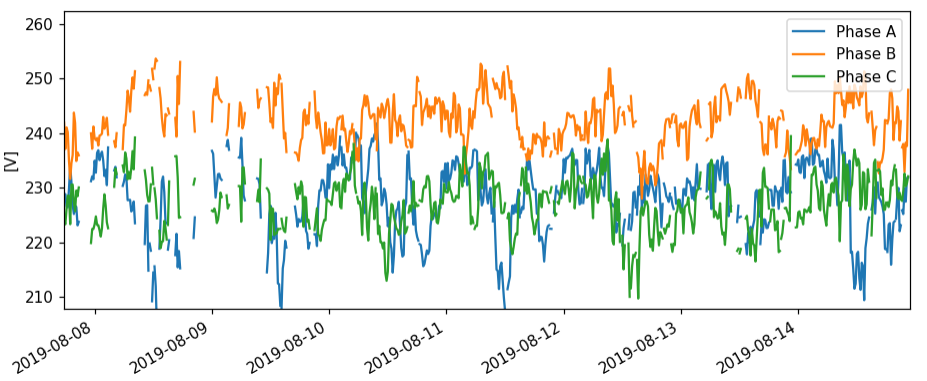}
     }
     \hfill
     \subfloat[Active power energy load at distribution substation. \label{fig:sample-load}]{%
       \includegraphics[width=0.5\textwidth]{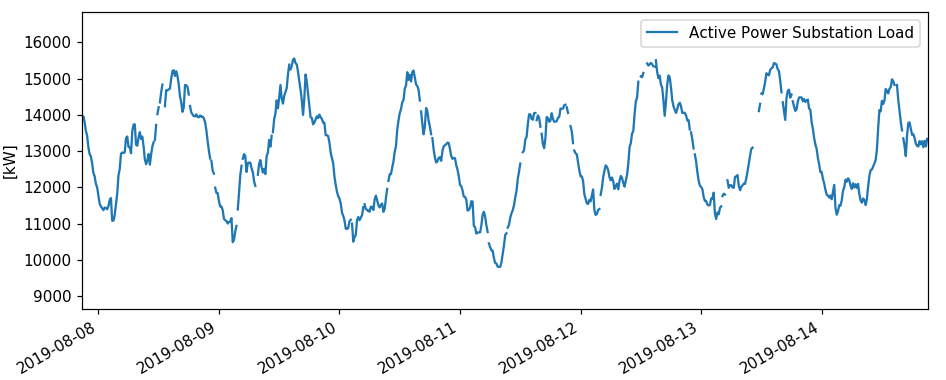}
     }
\caption{Example of available data}
\label{fig:sample-data}
\end{figure}

A smart-grid pilot deployment of an energy flexibility market, as part of the research project \cite{goflex}, was available at the Electricity Authority of Cyprus, the operator of the electrical distribution network in Cyprus. As also overviewed in Section \ref{sec:architecture}, the project sought to develop data analytic-services for predicting localised congestions on the grid, in the form of voltage violations due to excessive distributed renewable generation, and eventually prevent them by issuing bids for purchasing energy flexibility on the market \cite{Eck2019eenergy}. 
Live data were collected from July 2018 through to the end of 2019. Figure \ref{fig:data-ingestion} shows the rate of data ingestion from January through March 2019, where, on average, 15 million readings were received monthly (nearly 1.4K per hour) from about 500 sensors.  
Specifically, the received data consisted of voltage and energy profile data at single- and three-phase electrical prosumers (consumers and producers) collected from smart meters, as well as active power and current load at distribution substation and feeder heads obtained through the utility supervisory control and data acquisition (SCADA) system. Sensor data were received at different rates, both daily and hourly, as it is also clear from the ingestion patterns appearing in Fig. \ref{fig:data-ingestion}. Resampling and integration was applied to obtain 15-minute energy load time series data, as shown in Fig. \ref{fig:sample-load}, from the raw active power data at the distribution feeder and transformers, sampled at 1-minute resolution. Interpolation was used to derive 15-minute voltage time-series data at the prosumers, as in the example shown in Fig. \ref{fig:sample-volt}, from the irregularly sampled raw observations \cite{Eck2019}.

\begin{figure}
\centering
\includegraphics[width=0.5\textwidth]{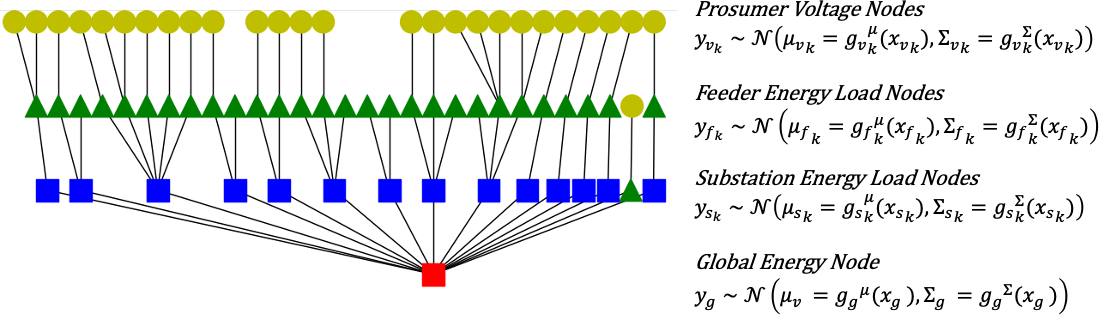}
\caption{Structure of proposed GNN model for the electrical grid. Shown are the decoding functions (\ref{eq:my_decoder}) for the  mean and variance at the voltage nodes (electrical prosumers) and at the energy nodes (feeders, substations, global grid node). The model is completed by encoding functions at the nodes, defined as in (\ref{eq:my_encoder}), and by message-passing functions  (\ref{eq:gnn_edge})-(\ref{eq:gnn_agg}), at the edges.}
\label{fig:graph}
\end{figure}

Ultimately, time-series were available for the energy load at 15 substations and 25 corresponding feeder heads, and for the voltage at 28 prosumers, 7 of which were 3-phase prosumers thus yielding a total of 48 voltage time-series. Historical temperature and solar irradiance observations at 15-minute resolution were also collected at each of the distribution substations, through a high-resolution weather data service provided by The Weather Company \cite{TWC2018}. Based on grid topology information, the GNN model structure as shown in Figure \ref{fig:graph} was derived.  Node encoding and decoding functions (\ref{eq:my_encoder}), (\ref{eq:my_decoder}) are defined to model sensor data and features relevant for each specific type of node. Temporal, autoregressive features for each time-series at 24-, 36- and 48-hour lags were considered to account for typical short-term seasonalities in electrical power consumption and generation patterns. Temperature and solar irradiance weather features at the substations nodes, along with autoregressive features at the same lags, are also included based on the availability of the weather data. 
As a result, the voltage mean and variance is modelled at the 28 prosumer nodes of dimensionality 8 (24) for single-phase (three-phase) prosumers. The mean and variance of the energy load at the 25 feeders and 15 substations is modelled with feeder nodes of dimensionality 8 and substation nodes of dimensionality 24 (including the weather features). An additional global node was included, with dimensionality 8 representing the mean and variance of the aggregated energy load for the whole DSO pilot.
The GNN model architecture is completed by the message-passing functions along the edges, namely (\ref{eq:gnn_edge})-(\ref{eq:gnn_agg}), which are defined to reflect the grid topology hierarchy going from the global node down to the substations, through their feeders and finally to the prosumers. All encoding, decoding and edge functions are implemented as multi-layer feedforward neural networks. As shown later, one of the benefits of incorporating domain knowledge in the proposed structure of the GNN is that it provides a significant reduction (more than $80\%$) in the parameter space of a centralised neural network model with comparable accuracy. 

The main use-case for the GNN model was the voltage prediction problem, which corresponds to the data imputation problem for the voltage time-series data at a given time $t$. It can be assumed that all autoregressive features, as well as weather and energy data at time $t$ are available. 

\subsection{Experimental Set-up and Benchmarks} \label{sec:results_benchmarks}

The GNN, designed as discussed in Section \ref{sec:results_data}, was trained on 1 year of historical data, from July 2018 to the end of June 2019, using the the negative log-likelihood objective function defined in  (\ref{eq:nll}). In order to avoid periods of excessive missing data, only samples with at most $10\%$ missing data were included in the training set, resulting in 27137 training samples. The training set was augmented with the same number of samples were the voltage is missing, in order to improve the performance on the voltage prediction task, as suggested in section \ref{sec:model_missingdata}. A gradient-descent training algorithm with a batch size limited to 5000 samples and based on the Adam optimizer \cite{Kingma2015}, with 0.01 learning rate, was utilised. Early stopping was employed to avoid overfitting based on results obtained on a 1-month cross-validation period, in July 2019. Different architecture choices for the number of layers of the encoding/decoding nodes and edge functions, for the size of the latent space at the nodes and for the number of message-passing iterations, were compared. A hyperbolic-tangent activation function was used throughout the experiments as found to give best convergence properties. Final test results were evaluated on data collected in August 2019. 
Centralised models based on multi-layer feed-forward neural networks (MLP) and auto-encoding (AE) neural networks were used as reference benchmarks. The same activation function, gradient-descent algorithm and convergence criteria as for the GNN model were used for both model classes. Also, the same set of features and, specifically for the AE, the same overall latent-space dimension as for the GNN model were used. While alternative, more complex neural-network architectures exist, the proposed choice for the benchmarks reflects the current state-of-the art in power systems modelling \cite{Barbeiro2014,Zamzam2019} and, at the same time, is considered sufficient to support the main conclusions of this work. 

\subsection{Voltage Prediction} \label{sec:results_voltage}
 
The GNN grid model, designed as in section \ref{sec:results_data}, was applied to the voltage prediction problem, as in the example of Figure \ref{fig:prediction-detail}. As discussed in section \ref{sec:model_missingdata}, variable prediction using the proposed GNN model corresponds to a data imputation problem, where inference on the graph is run while assuming that the voltage observations at each sample $t$ are unknown. From an initialization of the unknown data with a placeholder value, the GNN inference is iteratively run with the most recent estimate for the unknowns \cite{Boquet2019}. It was observed that after $5$ inference steps, the data imputation procedure consistently converged to negligible updates for the estimate of the unknowns. 

Different architectures for the GNN, structured as in Fig. \ref{fig:graph}, were explored by varying the number of layers for the MLPs implementing the node and edge functions, and the number of message-passing (MP) iteration steps.  The dimension of the latent space was fixed to $6$ for the voltage and feeder load nodes (with dimensionality $8$) and to $24$ for the substation nodes and three-phase voltage nodes (with dimensionality $24$), as no significant gains were obtained with increasing values. This finding is consistent with practical domain intuition: energy and voltage data are correlated with the 3 lagged features; similarly, some cross-correlation at voltage between different phases and between weather and energy load allows for further compression in latent space of substation load nodes and three-phase prosumer nodes. 

\begin{table}[ht]
    \centering
    \begin{tabular}{l||ccc|ccc}
    \hline 
    \rule{0pt}{2ex} 
    Model &\# Layers &\# MP &\# Params & MAPE & RMSE \\
    \hline 
     GNN &$2$  &$5$ &$143,464$ &$0.81\%$ & $2.42$ \\
     GNN &$3$  &$5$ &$169,144$  &$0.82\%$&  $2.45$ \\
     GNN &$2$  &$8$ &$142,832$ &$0.80\%$ &$2.40$\\
     GNN &$3$  &$8$ &$169,144$  &$0.81\%$&  $2.41$ \\
     MLP &$2$   &$-$ & $1,271,440$ & $0.80\%$ & $2.40$\\
     MLP &$3$   &$-$ & $2,118,760$ & $0.73\%$ & $2.21$\\
    AE &$2$ &$-$ & $2,407,088$ & $0.72\%$ & $2.17$ \\
    AE &$3$ &$-$ & $3,796,840$ & $0.70\%$ & $2.13$ \\
    \hline
    \end{tabular}
    \vspace{0.2cm}        
    \caption{Summary test results for GNN, MLP and AE models. Number of layers represent encoding/decoding and edge functions in GNNs and AE. MP denotes message-passing steps in GNNs. }
    \label{tab:results_voltage}
\end{table}

Table \ref{tab:results_voltage} reports the error metrics, in terms of mean-absolute-percentage error (MAPE) and root-mean-square error (RMSE), for different choices of the hyper-parameters in the GNN and benchmark models (MLP and AE). It is interesting to see how a comparable accuracy is obtained with respect to the MLP and AE models, but at a more than $80\%$ reduction in the number of neural-network weights. By exploiting sparsity conditions that are readily available from basic types of domain knowledge, such as grid connectivity in this case, GNNs allow for much more efficient machine learning models without sacrificing modelling accuracy. While it is possible to enforcing sparsity in traditional MLP or AE models, for example based on heuristic search over the layers connectivity space, the computational complexity makes this often prohibitive in practice. 

\begin{figure}
\centering
\includegraphics[width=0.5\textwidth]{prediction-detail}
\caption{Probabilistic voltage magnitude prediction at one of the prosumers. The shaded area is the  $95.4\%$ confidence interval (2 standard deviations) around the predicted mean. }
\label{fig:prediction-detail}
\end{figure}

\begin{table}[ht]
    \centering
    \begin{tabular}{l||ccccc}
    \hline 
    \rule{0pt}{2ex} 
    Missing data [$\%$] &$0$ &$0.1$ &$1$ &$5$ &$10$ \\
    \hline 
    Samples &$187$ &$356$ &$1108$  &$1472$ &$192$  \\
    MAPE [$\%$] & $0.810$ &$0.817$ &$0.815$ &$0.824$ &$0.854$\\
    \hline
    \end{tabular}
    \vspace{0.2cm}    
    \caption{Impact of missing data on MAPE of GNN model from Table \ref{tab:results_voltage} with 2-layer MLP functions and 5 message-passing steps. }
    \label{tab:results_voltage_missing}
\end{table}

The live data feed during the course of the demonstration project was affected by instances of missing data, often at random, as it also appears from some gaps in the actual data of Fig. \ref{fig:prediction-detail}. The performance of the proposed GNN model on the voltage prediction task was also evaluated with respect to the rate of missing data. Note that since, based on our formulation, the prediction problem is treated as a data imputation problem, no further change to the inference algorithm is required in order to handle missing values for the features. As summarized in Table \ref{tab:results_voltage_missing}, most of the samples in the test period (August 2019) had a missing data rate of $0.1-1\%$ (1108 samples) and $1-5\%$ (1472). While a minor degradation is observed with respect to the $187$ samples with no missing data, the performance remains fairly stable even for instances of $5-10\%$ missing data ratio. 

\subsection{Congestions and market bidding} \label{sec:results_bidding}

\begin{figure}[t]
\centering
\includegraphics[width=0.5\textwidth]{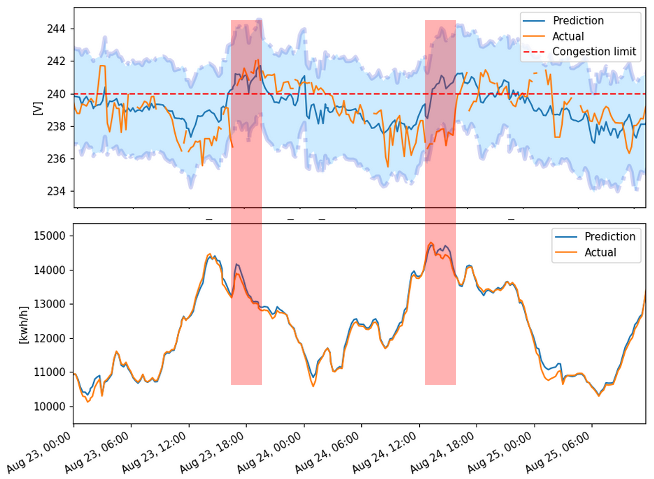}
\caption{Example of voltage congestion prediction service (top) and market bidding service, based on the estimate of energy flexibility (bottom) required to avoid the congestion. }
\label{fig:example-bidding}
\end{figure}

The same energy system GNN model, validated on the voltage prediction problem in section \ref{sec:results_voltage}, was utilised to inform congestion prediction and bidding data services, as outlined in section \ref{sec:architecture}. The probabilistic voltage predictions generated by the model were used to design statistical tests for determining the likelihood of an occurrence of a voltage congestion based on user-defined criteria. As an example, Fig. \ref{fig:example-bidding} (top) shows examples of congestions flagged as voltage exceeding a threshold of $240\,V$, based on a Z-test with probability 0.8413 (mean estimate exceeds the threshold by 1 standard deviation). The amount of energy variation required at a given point of the grid in order to avoid the congestion is then estimated by running inference on the GNN model with the affected voltage variables set to the desired maximum value and with the energy variables at the corresponding substation and feeder nodes treated as missing points. Figure \ref{fig:example-bidding} (bottom) shows a comparison between the resulting estimate of the energy load and its actual value, their difference representing the amount of energy variation required to  maintain the voltage below the desired threshold.  Based on this difference, a bid on an energy flexibility market can be placed by the grid operator to purchase  a corresponding increase or reduction in energy. 

Unlike for the voltage prediction service, a quantitative validation of the bidding service was not available based on the smart-grid pilot demonstration set-up. Specifically, a ground-truth value for the estimated energy variation can not be physically measured and the amount of energy flexibility available within the market was too small to generate variations that are visible in the data. It can be, however, noted, based on the example in Fig. \ref{fig:example-bidding}, how the model suggests increasing the energy load (increase demand or reduce generation) to reduce the voltage profile, which is qualitatively consistent with the physics of power flow models.

\section{Conclusion} \label{sec:conclusion}
A novel probabilistic graphical model for energy systems, based on the framework of graph neural networks (GNNs), was introduced. The model was used to inform several data-analytics services for electrical system operators, such as the prediction of grid congestions and the estimation of the amount of energy flexibility required in order to avoid such congestions. Both problems are treated as data imputation inference on the graph. The model was evaluated in the context of a smart-grid demonstration project, to predict and manage voltage congestions due to high levels of distributed solar generation. 

With respect to existing probabilistic graphical models, the proposed approach can learn the belief propagation algorithm from the data, instead of relying on ad-hoc, heuristics-based rules that are hard to generalise. Furthermore, the sparsity in the GNN, informed by the domain knowledge, significantly reduces the number of model parameters when compared with traditional machine-learning models with similar accuracy. For example, when compared with a traditional MLP model, a reduction of more than $80\%$ in the number of neural-network weights is achieved. 

The proposed GNN model can also efficiently adapt to changes in the electrical systems or in the available sensor data, by adapting the graph structure accordingly and requiring retraining only on the parameters of the affected portions of the graph. These topics will be scope for future research.


\section*{Acknowledgement}
The authors wish to acknowledge Ioannis Papageorgiou from the Electricity Authority of Cyprus for providing access to and understanding of the data used in this study.
This research has received funding from the European
Research Council under the European Unions Horizon 2020
research and innovation programme (grant agreement no.
731232).



\bibliographystyle{IEEEtran}
\balance
\bibliography{mybib}

\begin{thebibliography}{10}
\providecommand{\url}[1]{#1}
\csname url@samestyle\endcsname
\providecommand{\newblock}{\relax}
\providecommand{\bibinfo}[2]{#2}
\providecommand{\BIBentrySTDinterwordspacing}{\spaceskip=0pt\relax}
\providecommand{\BIBentryALTinterwordstretchfactor}{4}
\providecommand{\BIBentryALTinterwordspacing}{\spaceskip=\fontdimen2\font plus
\BIBentryALTinterwordstretchfactor\fontdimen3\font minus
  \fontdimen4\font\relax}
\providecommand{\BIBforeignlanguage}[2]{{%
\expandafter\ifx\csname l@#1\endcsname\relax
\typeout{** WARNING: IEEEtran.bst: No hyphenation pattern has been}%
\typeout{** loaded for the language `#1'. Using the pattern for}%
\typeout{** the default language instead.}%
\else
\language=\csname l@#1\endcsname
\fi
#2}}
\providecommand{\BIBdecl}{\relax}
\BIBdecl

\bibitem{Child2019}
M.~Child, C.~Kemfert, D.~Bogdanov, and C.~Breyer, ``Flexible electricity
  generation, grid exchange and storage for the transition to a 100
  energy system in europe,'' \emph{Renewable Energy}, vol. 139, pp. 80--101,
  2019.

\bibitem{Abur2004}
A.~Abur and A.~G. Exposito, \emph{{Power System State Estimation: Theory and
  Implementation}}.\hskip 1em plus 0.5em minus 0.4em\relax CRC Press, 2004.

\bibitem{Monticelli2000}
A.~Monticelli, ``{Electric power system state estimation},'' \emph{Proceedings
  of the IEEE}, vol.~88, no.~2, pp. 262--282, feb 2000.

\bibitem{Barbeiro2014}
P.~N.~P. Barbeiro, J.~Krstulovic, H.~Teixeira, J.~Pereira, F.~J. Soares, and
  J.~P. Iria, ``State estimation in distribution smart grids using
  autoencoders,'' in \emph{Proceedings of the {IEEE} 8th International Power
  Engineering and Optimization Conference ({PEOCO})}, Langkawi, Malaysia, 2014.

\bibitem{Zamzam2019}
A.~S. Zamzam, X.~Fu, and N.~D. Sidiropoulos, ``{Data-Driven Learning-Based
  Optimization for Distribution System State Estimation},'' \emph{IEEE
  Transactions on Power Systems}, vol.~34, no.~6, pp. 4796--4805, apr 2019.

\bibitem{Zhang2019}
L.~Zhang, G.~Wang, and G.~B. Giannakis, ``{Real-Time Power System State
  Estimation andForecasting via Deep Unrolled Neural Networks},'' \emph{IEEE
  Transactions on Signal Processing}, vol.~67, no.~15, pp. 4069--4077, aug
  2019.

\bibitem{Schumilin2017}
A.~Schumilin, K.-U. Stucky, F.~Sinn, and V.~Hagenmeyer, ``Towards
  ontology-based network model management and data integration for smart
  grids,'' in \emph{In Proc of the Workshop on Modeling and Simulation of
  Cyber-Physical Energy Systems (MSCPES), Pittsburgh, PA, USA}, 2017.

\bibitem{Zhang2017}
Z.~J. Zhang, ``Graph databases for knowledge management,'' \emph{{IEEE} IT
  Pro}, vol.~19, no.~6, pp. 26 -- 32, 2017.

\bibitem{Eck2019}
B.~Eck, F.~Fusco, R.~Gormaly, M.~Purcell, and S.~Tirupathi, ``Scalable
  deployment of {AI} time-series models for {IoT},'' in \emph{Workshop AI for
  Internet of Things (AI4IoT) at the 28th International Joint Conference on
  Artificial Intelligence (IJCAI)}, 2019.

\bibitem{Chen2018}
B.~Chen, B.~Eck, F.~Fusco, R.~Gormally, M.~Purcell, M.~Sinn, and S.~Tirupathi,
  ``Castor: Contextual {IoT} time series data and model management at scale,''
  \emph{Proc. of the 18th ICDM 2018, pp 1487-1492}, 2018.

\bibitem{Ploennigs2017}
J.~Ploennigs, A.~Ba, and M.~Barry, ``Materializing the promises of cognitive
  {IoT}: How cognitive buildings are shaping the way,'' \emph{{IEEE} Internet
  of Things Journal}, 2017.

\bibitem{Hu2011}
Hu, Y., Kuh, A., Kavcic, A., Yang, and T., ``{A Belief Propagation Based Power
  Distribution System State Estimator},'' \emph{IEEE Computational Intelligence
  Magazine}, no. AUGUST, pp. 36--46, 2011.

\bibitem{Cosovic2016a}
M.~Cosovic and D.~Vukobratovic, ``Distributed gauss-newton method for ac state
  estimation: A belief propagation approach,'' \emph{IEEE Int. Conf. on Smart
  Grid Comm.}, 2016.

\bibitem{Fusco2018}
F.~Fusco, ``Probabilistic graphs for sensor data-driven modelling of power
  systems at scale,'' in \emph{Data Analytics for Renewable Energy Integration.
  Technologies, Systems and Society - 6th {ECML} {PKDD} Workshop, {DARE} 2018,
  Dublin, Ireland, September 10, 2018, Revised Selected Papers}, 2018.

\bibitem{Scarselli2009}
F.~Scarselli, M.~Gori, A.~C. Tsoi, M.~Hagenbuchner, and G.~Monfardini, ``The
  graph neural network model,'' \emph{IEEE Transactions on Neural Networks},
  vol.~20, no.~1, pp. 61--80, 2009.

\bibitem{Casas2020}
S.~Casas, C.~Gulino, R.~Liao, and R.~Urtasun, ``Spagnn: Spatially-aware graph
  neural networks for relational behavior forecasting from sensor data,'' in
  \emph{Proceedings of the International Conference on Robotics and Automation
  (ICRA)}, 2020.

\bibitem{goflex}
GOFLEX, ``Generalized operational flexibility for integrating renewables in the
  distribution grid,'' \url{https://goflex-project.eu/}, Accessed:2020-05-23.

\bibitem{Eck2019eenergy}
B.~Eck, F.~Fusco, R.~Gormally, M.~Purcell, and S.~Tirupathi, ``{AI} modelling
  and time-series forecasting systems for trading energy flexibility in
  distribution grids,'' in \emph{e-{Energy} '19}.\hskip 1em plus 0.5em minus
  0.4em\relax New York, NY, USA: Association for Computing Machinery, 2019, p.
  381–382.

\bibitem{Koller2009}
D.~Koller and N.~Friedman, \emph{Probabilistic Graphical Models}.\hskip 1em
  plus 0.5em minus 0.4em\relax MIT Press, 2009.

\bibitem{Loelinger2007}
H.-A. Loelinger, J.~Dauwels, J.~Hu, S.~Korl, L.~Ping, and F.~R. Kschischang,
  ``The factor graph approach to model-based signal processing,''
  \emph{Proceedings of the IEEE}, 2007.

\bibitem{Battaglia2018}
P.~W. Battaglia, J.~B. Hamrick, V.~Bapst, A.~Sanchez{-}Gonzalez, V.~F.
  Zambaldi, M.~Malinowski, A.~Tacchetti, D.~Raposo, A.~Santoro, R.~Faulkner,
  {\c{C}}.~G{\"{u}}l{\c{c}}ehre, H.~F. Song, A.~J. Ballard, J.~Gilmer, G.~E.
  Dahl, A.~Vaswani, K.~R. Allen, C.~Nash, V.~Langston, C.~Dyer, N.~Heess,
  D.~Wierstra, P.~Kohli, M.~Botvinick, O.~Vinyals, Y.~Li, and R.~Pascanu,
  ``Relational inductive biases, deep learning, and graph networks,''
  \emph{arXiv}, vol. abs/1806.01261, 2018.

\bibitem{Gilmer2017}
J.~Gilmer, S.~S. Schoenholz, P.~F. Riley, O.~Vinyals, and G.~E. Dahl, ``Neural
  message passing for quantum chemistry,'' in \emph{Proceedings of the 34th
  International Conference on Machine Learning, Sydney, Australia}, 2017.

\bibitem{Boquet2019}
G.~{Boquet}, J.~L. {Vicario}, A.~{Morell}, and J.~{Serrano}, ``Missing data in
  traffic estimation: A variational autoencoder imputation method,'' in
  \emph{Proceedings of the EEE International Conference on Acoustics, Speech
  and Signal Processing (ICASSP)}, 2019, pp. 2882--2886.

\bibitem{TWC2018}
{The Weather Company}, ``Cleaned observations api,''
  http://cleanedobservations.wsi.com/documents/
  WSI
  Accessed:2020-05-23.

\bibitem{Kingma2015}
D.~P. Kingma and J.~Ba, ``Adam: A method for stochastic optimization,''
  \emph{CoRR}, vol. abs/1412.6980, 2015.

\end{thebibliography}
%

\end{document}